# Addressing Class Variable Imbalance in Federated Semi-Supervised Learning

Zehui Dong, Wenjing Liu , Siyuan Liu and Xingzhi Chen

School of Data Science and Application, Inner Mongolia University of Technology, Hohhot, China

## Abstract

*Federated Semi-supervised Learning (FSSL) combines techniques from both fields of federated and semi-supervised learning to improve the accuracy and performance of models in a distributed environment by using a small fraction of labeled data and a large amount of unlabeled data. Without the need to centralize all data in one place for training, it collect updates of model training after devices train models at local, and thus can protect the privacy of user data. However, during the federal training process, some of the devices fail to collect enough data for local training, while new devices will be included to the group training. This leads to an unbalanced global data distribution and thus affect the performance of the global model training. Most of the current research is focusing on class imbalance with a fixed number of classes, while little attention is paid to data imbalance with a variable number of classes. Therefore, in this paper, we propose Federated Semi-supervised Learning for Class Variable Imbalance (FCVI) to solve class variable imbalance. The class-variable learning algorithm is used to mitigate the data imbalance due to changes of the number of classes. Our scheme is proved to be significantly better than baseline methods, while maintaining client privacy.*

## 1. Introduction

In recent years, the rapid development of both Internet of Things (IoT) and Artificial Intelligence (AI) technologies has driven the boom of mobile smart devices, but also the massive amount of data generation that accompanies them. According to Cisco (Cisco), it is expected that by 2025, the data generated by IoT devices worldwide at the edge of the network will reach 79.4 ZB [1]. Faced with the data explosion brought about by the rapid development, the cloud computing model converges IoT and AI by using the data generated by IoT devices to train machine learning models and then using the models to analyze the data (e.g., image recognition, intelligent prediction, etc.), which in turn drives the implementation of AI and IoT in various application scenarios [2]. However, this model requires all data to be aggregated to a cloud data center and then processed centrally, which can lead to huge transmission delays and privacy leaks.

The combination of Edge Computing (EC) [3] and Federated Learning (FL) [4] offers the possibility to solve the above problems. First, EC runs deep learning model training tasks on edge nodes close to data sources by shifting from centralized to distributed edge networks based on cloud data centers. Second, FL unites many edge nodes at the edge of the network to jointly participate in distributed collaborative model training on the basis of not sharing data. In this way, not only the latency of the data transmission process is reduced, but also the level of security and privacy protection of the whole training process is improved.

However, at the data level, the data collected by different devices are scattered and variable due to the different geographical locations of the devices [5], and the data used for FL model training is class unbalanced. These class-imbalanced data are reflected in the model classification task as class imbalance [6], and the models trained using such data are limited. In the case of a binary classification task, the ratio between positive and negative data may be as high as 999:1, and it is difficult for a model trained with such data to identify the data classes that account for a relatively small amount of data [7]. Meanwhile, FL enables multiple devices to collaboratively train a global model by aggregating the local model parameters of multiple devices, without the need to upload data from local to the cloud. However, in FL iterative training, after some devices stop participating in the training midway, it will result in the loss of some data, as well as new devices joining the training will add new data, which in turn will lead to an unbalanced global data distribution, thus affecting the performance of the global model [8]. Most of the current research is focusing on class imbalance with a fixed number of classes [6], while little attention has been paid to class imbalance with a changing number of classes.

In edge networks, we found that for most mobile smart devices, most of their collected data are unlabeled data [9], and only a small fraction of them are labeled data, which are unable to support model training. A potential solution is to use Federated Semi-supervised Learning [10] (FSSL) approach, which updates FL models using labeled data and unlabeled data with pseudo-labels by using semi-supervised learning to train these data and generating pseudo-labels for unlabeled data, this approach is very effective in the case of class imbalance and lack of labels. Since FL training is performed by exchanging gradients in an encrypted form, the training data is not fully observable to the edge server and aggregation server. Therefore, it is crucial to detect the category changes in FL and mitigate their impact.

In order to better address the class imbalance problem of class number changes, this paper proposes Federated Semi-supervised Learning for Class Variable Imbalance (FCVI) to address class variable imbalance, which enables the model to be trained on a large amount of decentralized data, while can effectively solve the class imbalance problem in the dataset. The main contributions of this paper are 3-fold.

(1) A federal gradient monitoring scheme is designed, which can infer the changes of each class in the FL training process through the gradient without transmitting other data, ensuring the privacy protection of the model training process.
(2) A class-variable learning algorithm is designed. After the monitoring scheme monitors the class changes, we selectively expand the data using unlabeled data and semi-supervised learning self-training methods to alleviate the class imbalance due to the changes in the number of classes.
(3) A real scenario with imbalanced image data is deployed to design distributed training experiments using an image classification model, which in turn improves the accuracy of model recognition.

## 2. RELATED WORK

### 2.1. Class Imbalance Machine Learning

Research on class imbalance machine learning has focused on supervised learning and semi-supervised learning.

In Supervised Learning (SL), models require labeled data to update their parameters, and their approaches to address class imbalance can be divided into data-level and algorithm-level. The data level utilizes data oversampling or undersampling methods. In the literature [11], the authors use an oversampling approach to replicate randomly selected samples from a small number, which reduces both inter-category and intra-category imbalances, but such an approach can lead to the occurrence of overfitting. In the literature [12], the authors use the undersampling method to remove randomly from the majority of classes until all classes have the same number of samples, but the significant disadvantage of this is that it discards a portion of the available data. Unlike the data level, the algorithm level is to modify the training algorithm or the network structure. For example, in the literature [13], the authors use a re-weighting approach to assign weights to different training data and address the problem of low accuracy in a few classes by equalizing the loss function and ignoring the gradient of most classes. All the above methods are tuned with labeled data, while their performance is unknown in the absence of labeled data.

In semi-supervised learning (SSL), label-free data can be used to improve model performance. In the literature [14], the authors found that SSL using unlabeled data can mitigate the impact of data imbalance on model training. In the literature [15], the authors propose a method to suppress the loss of consistency to suppress the loss of a few classes and thus mitigate the impact of data imbalance. In the literature [16], the authors mitigate the data imbalance in SSL scenarios by softening the pseudo-labels generated in the deviation model through convex optimization. In the literature [17], the authors eliminate the effect of unbalanced data by pseudo-labeling unlabeled data and then expanding these data into labeled data to improve the distribution of the data.

### 2.2. Class Imbalance Federal Learning

Due to the high computational effort of training machine learning models, researchers have been exploring the use of multiple devices to learn a general model. Federated learning has emerged as an effective solution to learn a global model while keeping all training data on the local device. In federated learning, its research has focused on reducing communication overhead and preserving privacy, and only a few studies have paid attention to the problem of class imbalance in training data leading to a decrease in model accuracy. In the literature [18], the authors use weighted optimization to alleviate the imbalance problem by assigning higher weights to the nodes that produce better training results. However, this method makes the network bandwidth increase, which in turn increases the network burden, and uploading local data to the server causes privacy issues. In the literature [19], the authors dealt with this by reassigning training nodes and grouping them for training based on their local and global similarity degree. But it does not solve the problem of unbalanced local data distribution of nodes. In the literature [20], the authors improved the distribution of unbalanced data using data augmentation methods such as random displacement, random rotation, random shearing, and random scaling. Because the augmented data are similar data generated on the original data, this makes the model training more prone to overfitting.

# 3. DESIGN OF FCVI

## 3.1. Problem Definition

In the conventional training scenario, the model training is focused on a single device and the category distribution is known, while the training data are labeled and sufficient. In the edge network FSSL, the model training is distributed over an aggregation server and an edge server and the class distribution is not known, while the training data consists of a large amount of unlabeled data and a small amount of labeled data. Specifically, there is one label set $\mathcal{D}_i^{(L)} = \left\{ (x_j, y_j) : j \in (1, \cdots, N) \right\}$ on each edge server $i$. The number of training samples for class $l$ in label set $\mathcal{D}_i^{(L)}$ is denoted as $N_i^l$, i.e., $\sum_l^L N_l = N$. In addition to label set $\mathcal{D}_i^{(L)}$, there is an unlabeled set $\mathcal{D}_i^{(U)} = \{ u_c \in \mathbb{R}^d : c \in (1, \cdots, C) \}$, where $C \gg N$, label set $\mathcal{D}_i^{(L)}$ and unlabeled set $\mathcal{D}_i^{(U)}$ have the same class distribution. We use the label fraction $\beta = N / (N + C)$ to represent the percentage of labeled data to the total data.

In FSSL, local model training using labeled data can be considered as regular centralized learning, so the local model training is on a multilayer feedforward neural network with inputs in the feature space $X$ and label space $Y = \{1, \cdots, L\}$, and the $L$ output size of the classifier is equal to the number of classes. In this paper, all the intermediate layers are combined into a single hidden layer $HL$, containing $s$ neurons. We send the $j$-th sample of class $l$, denoted as $X_j^{(l)}$, to the classifier $L$, whose output corresponding to $HL$ is denoted as $Y_j^{(l)} = \left[ y_{j,(1)}^{(l)}, \cdots, y_{j,(s)}^{(l)} \right]$. the output of the last layer is $Z_j^{(l)} = \left[ z_{j,(1)}^{(l)}, \cdots, z_{j,(L)}^{(l)} \right]$, and then the $softmax$ operation is performed to obtain the probability vector $S = \left[ f_{j,(1)}^{(l)}, \cdots, f_{j,(L)}^{(l)} \right]$. the function $f : \{X \Rightarrow S\}$ will be mapped to the output probability simplex $S$. $f$ parameterizes the hypothetical class $\mathbb{W}$, i.e., the overall parameters of the classifier. In addition, the connection weights from $HL$ to the output layer are denoted as $W = \left[ W_{(1)}, W_{(2)}, \cdots, W_{(Q)} \right]$ and $W \in \mathbb{W}$. In each training iteration, we apply the backpropagation method to calculate the gradient of the loss function $L(\mathbb{W})$ under $\mathbb{W}$. Denoting the weight of the $t$-th training iteration by $\mathbb{W}(t)$ and the learning rate by $\lambda$, we then have $\mathbb{W}(t + 1) = \mathbb{W}(t) - \lambda \nabla L(\mathbb{W}(t))$.

## 3.2. Federal Gradient Monitoring Method

In edge intelligence scenarios, the data collected by edge devices are differentiated. Because of the different locations of the edge devices, the data collected by them are also different from other edge devices in terms of categories. At the same time, edge devices are often mobile and unstable, therefore, in the iterative training of the FL model, some of the devices will no longer participate in the training, which will result in the loss of some categories of data, and new devices will add new categories of data after joining the training, which will lead to changes in the number of data categories in the iterative training of the FL model, thus affecting the performance of the global model. In order to detect and mitigate the performance degradation caused by class-variable imbalance, we design a global monitoring method based on FL gradient to estimate the class variation of FL model iterative training as follows.

In any real-valued neural network $f$, when its last layer is a linear layer with $softmax$ operations, then for any input samples $X_i^{(p)}$ and $X_j^{(p)}$ of the same class $p$, if the inputs $Y_i$ and $Y_j$ in the last layer are the same, then the gradient $W$ of the connection weights caused by and between the layers before the last layer $X_i^{(p)}$ and $X_j^{(p)}$ is the same [21].

In small-batch (mini-batch) training, the gradients of the samples within a batch are accumulated to update the model parameters, i.e.

$$\Delta_{batch}W = -\frac{\lambda}{n^{batch}}\sum_{p=1}^{Q}\sum_{j=1}^{n^{(p)}}\nabla_{w_j^{(p)}}L(\mathbb{W}) \tag{1}$$

Data samples of the same class $p$ will have similar $Y^{(p)}$, and therefore the corresponding gradients are very similar. If the mean value of the gradient is $\overline{\nabla_{w^{(p)}}L(W)}$, then equation (1) can be written as

$$\Delta_{batch}W = -\frac{\lambda}{n^{batch}}\sum_{p=1}^{Q}(\overline{\nabla_{\nabla_{w^{(p)}}L(W)}} \cdot n^{(p)}) \tag{2}$$

where $n^{(p)}$ is the number of samples of class $p$ in the batch and $n^{batch}$ is the size of the batch. For a round of local training in FL, the total number of iterations (Iteration) of local gradient updates is $\left[\left(\sum_{p=1}^{Q}\frac{N_p}{n^{batch}}\right) \cdot N_{ep}\right]$, where $N_{ep}$ denotes the number of local Epochs. To illustrate the proportional relationship between the gradient size and the sample size, we assume that the parameters are relatively small and negligible within an Epoch. In this case, the different batches of $\overline{\nabla_{w^{(p)}}L(W)}$ within an Epoch remain constant, and we can aggregate them to obtain an Epoch whose weights are updated as

$$\Delta_{epoch}W = -\frac{\lambda}{n^{batch}}\sum_{p=1}^{Q}(\overline{\nabla_{w^{(p)}}L(W)} \cdot N_p) \tag{3}$$

where $N_p$ is the total number of $p$ class samples.

In the standard FL distributed training setting, the global server generally aggregates the selected local gradients by the FedAvg algorithm as follows.

$$\nabla L(W)_{t+1}^{Avg} = \frac{1}{K}\sum_{j=1}^{K}\nabla L(W)_{t+1}^{j} \tag{4}$$

where $K$ represents the number of clients.

Based on the above analysis, for any local training starting from the same current global model, the data samples of the same class $p$ output very similar $Y$ and similar $\overline{\nabla_{w^{(p)}}L(W)}$. In this case, the gradients obtained from class $p$ in a global epoch are

$$\Delta_{global}W^{(p)} = -\frac{\lambda}{n^{batch} \cdot K}\sum_{j=1}^{K}\left(\overline{\nabla_{w^{(p)}}^{j}L(W)} \cdot N_p^j\right)$$

$$= -\frac{\lambda}{n^{batch} \cdot K}\overline{\nabla_{w^{(p)}}L(W)}(\sum_{j=1}^{K}N_p^j) \tag{5}$$

Based on this relationship, we develop the following global monitoring method based on FL gradient. At round $t + 1$, when the aggregation server monitors the change in the number $K_{t+1}$ volume of the client, the global gradient by comparing the class $p$ at round $t$ is

$$\frac{\Delta_{global} W_t^{(p)}}{\Delta_{global} W_{t+1}^{(p)}} = \frac{-\frac{\lambda}{n^{batch} \cdot K_t} \overline{V_{w^{(p)}} L(W)} (\sum_{j=1}^{K_t} N_p^j)}{-\frac{\lambda}{n^{batch} \cdot K_{t+1}} \overline{V_{w^{(p)}} L(W)} (\sum_{j=1}^{K_{t+1}} N_p^j)}$$

$$= \frac{K_{t+1}}{K_t} \cdot \frac{\sum_{j=1}^{K_t} N_p^j}{\sum_{j=1}^{K_{t+1}} N_p^j} \qquad (6)$$

Because the aggregation server knows the number of edge servers $K_{t+1}$ in round $t + 1$ and the number of edge servers $K_t$ in round $t$, the change ratio $R_p$ of class $p$ samples can be obtained from Equation (6) as :

$$R_p = \frac{K_{t+1}}{K_t} \cdot \frac{\Delta_{global} W_{t+1}^{(p)}}{\Delta_{global} W_t^{(p)}} = \frac{\sum_{j=1}^{K_{t+1}} N_p^j}{\sum_{j=1}^{K_t} N_p^j} \qquad (7)$$

In the global monitoring method based on FL gradient, when the number of edge servers participating in FL training changes, the aggregation server uses Equation (7) after calculating the local FL gradient, we can obtain the change ratio vector $R = [R_1, \cdots, R_p, \cdots R_Q]$ for all classes in the current training round.

### 3.3. Class Variable Learning Algorithm

Once the aggregation server monitors a change in the number of edge servers, it indicates a change in the local and global data class distribution, which will negatively affect the performance of FL model training. Due to the different physical locations where the edge servers are located, some edge servers will have their class data samples that distinguish them from other edge servers. Therefore, among the class-variable imbalances caused by equipment movement, four categories can be classified according to the value of the change ratio $R_p$.

(1) When $R_p > 1$, then $\sum_{j=1}^{K_{t+1}} N_p^j > \sum_{j=1}^{K_{t+1}} N_p^j$, it is known that the global data class $p$ samples are increasing in round $t + 1$.
(2) When $0 < R_P < 1$, then $\sum_{j=1}^{K_{t+1}} N_p^j < \sum_{j=1}^{K_{t+1}} N_p^j$, it can be known that the $t + 1$ round of global data $p$ class samples are decreasing.
(3) When $R_p$ is an outlier, then $\sum_{j=1}^{K_t} N_p^j = 0$, it can be known that the number of $p$ class samples in the $t$ round of global data is 0.
(4) When $R_p = 0$ , then $\sum_{j=1}^{K_{t+1}} N_p^j = 0$, it is known that the number of $p$ class samples in the $t + 1$ round of global data is 0.

Single application of existing methods through local or global cannot effectively mitigate the impact of changes in data class distribution, so this paper designs a class-variable learning algorithm to mitigate the negative impact on FL model training by taking different approaches from local and global through the change ratio $R_p$ obtained from FL gradient monitoring method, respectively, and Algorithm 1 shows the details of the class-variable learning algorithm.

**Algorithm 1**:class-variable learning algorithm

**AggregationServer**:

1: Initialize parameters $\theta_0$;

2:    **for** each global epoch $t + 1$ in $1, \cdots, T$ **do**

3:        **for** each client $i$ in $1, \cdots, I_{t+1}$ **do**  //Collect edge server parameters

4:            receive edge server  $\theta_{i,t+1}$ ;

5:        **end for**

6:        $\theta_{t+1} = \sum_{i=1}^{|I_{t+1}|} \omega_i \theta_{i,t+1}$;  //Aggregate Edge Server Parameters

7:        **if** $I_{t+1} \neq I_t$ **then**  //Determine if the number of edge servers has changed

8:            **for** each class $l$ in $1, \cdots, L$ **do**

9:                $R^{(l)} \leftarrow$ FLGradientMonitorMethod($\theta_{t+1}^{(l)}, \theta_t^{(l)}$);

10:                **if** $R^{(l)} > 0$ **then**

11:                    $R_{min} = min(R_{min}, R^{(l)})$;

12:                **end if**

13:            **end for**

14:            **for** each $R^{(l)}$ in $1, \cdots, R^{(L)}$ **do**

15:                **if** $R^{(l)} > 0$ **then**

16:$\mu_{t+1}^{(l)} = \frac{R_{min}}{R^{(l)}}$;

17:                **else if** $R^{(l)} = 0$ **then**

18:                    $\theta_{t+1}^{(l)} = \theta_t^{(l)}$;

19:                    $\mu_{t+1}^{(l)} = 1$;

20:                **else**

21:                    $\mu_{t+1}^{(l)} = 1$;

22:                **end if**

23:            **end for**

24:            $\mu_{t+1} = \left[ \mu_{t+1}^{(1)}, \cdots \mu_{t+1}^{(l)}, \cdots \mu_{t+1}^{(L)} \right]$;

25:            $\theta_{t+1} = \left[ \theta_{t+1}^{(1)}, \cdots \theta_{t+1}^{(l)}, \cdots \theta_{t+1}^{(L)} \right]$;

26:        **end if**

**Edge Server** :

1: **for** each edge server $i$ in $1, \cdots, I_{t+1}$ **do**

2:    **for** each local epoch $t + 1$ in $1, \cdots, T$ **do**

3:    Class VariableSelf-trainingMethod($\mu_{t+1}$)

4:    **end for**

5:    **return** $\theta_{i,t+1}$ to AggregationServer ;

6: **end for**

First, when $R_p$ is (1), the number of class $l$ samples in the global data of round $t$ is 0, while the number of class $l$ samples in round $t + 1$ is not 0, indicating that the number of categories increases, and the aggregation server aggregates the class parameters normally in round t1 without performing other operations. Secondly, when $R_p$ is (2), the number of class $l$ samples in the global data in round $t$ is not 0,while the number of class $l$ samples in the $t + 1$ round is 0, indicating that the number of categories decreases,and the aggregation server uses round $t$ to update the aggregation of the class parameters in round $t + 1$ to avoid the problem of gradient disappearance due to the number of samples in the class being 0.Finally, when $R_p$ is (3) or (4), selective data expansion is performed on the local edge server using unlabeled data through the SSL self-training method.

The self-training method [22] is a widely used iterative approach in semi-supervised learning. Traditionally, self-training involves training an initial model by using a small amount of labeled data, then using that model to predict unlabeled samples, and then selecting unlabeled data with high confidence to expand the label set. This process actually involves the classification model using its own prediction results to improve the model performance. The self-training approach requires only an initial model, a small amount of labeled data, and a large amount of unlabeled data to perform a complex semi-supervised learning task. Specifically, self-training works as follows.

(1) Step 1: An initial model is first trained using the labeled set $\mathcal{D}_i^{(L)} = \{(x_j, y_j): j \in (1, \cdots, N_i)\}$.

(2) Step 2: The trained classification model is used to predict the class labels of all the unlabeled data $u_c$. Among these predicted class labels, the one with the highest correct rate and exceeding the probability threshold is considered as the pseudo-label $\hat{y}_c$, and the set of unlabeled data with such labels is the pseudo-label set $\widehat{\mathcal{D}}_i^{(U)} = \{(u_c, \hat{y}_c): c \in (1, \cdots, C_i)\}$.

(3) Step 3: Include the pseudo-label set $\widehat{\mathcal{D}}_i^{(U)}$ in the label set $\mathcal{D}_i^{(L)}$, i.e., $\acute{\mathcal{D}}_i^{(L)} = \mathcal{D}_i^{(L)} \cup \widehat{\mathcal{D}}_i^{(U)}$. Retrain the model on the new label set.

(4) Iterate step 2 and step 3 repeatedly until the predicted class labels in step 2 are not satisfying the probability threshold or until no unlabeled data are retained.

To accommodate the class imbalance, we modify the self-training method. Instead of including all the data of pseudo-label set $\widehat{\mathcal{D}}_i^{(U)}$ in the label set $\mathcal{D}_i^{(L)}$, we select a subset $\acute{S}_i^{(U)}(\acute{S}_i^{(U)} \subset \widehat{\mathcal{D}}_i^{(U)})$ from it to expand the label set $\widehat{\mathcal{D}}_i^{(U)}$, i.e., $\acute{\mathcal{D}}_i^{(L)} = \mathcal{D}_i^{(L)} \cup \acute{S}_i^{(U)}$. The selection rule for this subset $\hat{S}_i^{(U)}$: the smaller the change ratio $R_p$ of class $p$, the more unlabeled data are predicted as class $p$. Specifically: the amount of sample change $R_p$ of class $p$ is known, and the unlabeled data corresponding to being predicted as class $p$ are contained in a pseudo-labeled subset ratio of

$$\mu_p = \frac{R_{min}}{R_p}$$

where $R_{min}$ is the minimum value of the amount of sample change in class $p$. For example, when class $p$ is the class with the smallest amount of change in each class ($R_{min} = R_p$) and the unlabeled data of class $p$ is included in the pseudo-labeled subset ratio $\mu_p = 1$, then all the unlabeled data of class $p$ is retained. When $R_{min} = \frac{1}{2}$, $R_p = 2$, the unlabeled data of class $p$ is contained in the pseudo-labeled subset ratio $\mu_p = \frac{1}{4}$, then the unlabeled data of $\frac{1}{4}$ in class $p$ is retained.

## 4. EXPERIMENTS AND RESULTS

In this paper, the Convolutional Architecture for Fast Feature Embedding (Caffe) [23] deep learning framework is used to implement the proposed multilevel class rebalancing distributed method and its accuracy is verified by comparison.

### 4.1. Experimental Configuration

#### 4.1.1. Dataset and Model

Simulating the classification task in a real environment, the ImageNet dataset is chosen to construct the class-variable imbalance dataset in this paper. The ImageNet dataset is a large image classification dataset established to promote the development of computer graphics recognition technology, with a large number and high resolution of thousands of classes of image data. We used the ILSVRC2012 dataset, which contains 1000 categories and 1.2 million images. We selected 15 categories out of the 1000 categories, each containing 1300 images, and then divided them into 15600 training images and 3900 test images according to the allocation ratio of 8:2 between the training and test sets. For the training images, labeled data accounts for 30% of all data, and the remaining 70% of data is unlabeled data, i.e., label score $\beta = 0.3$.

To observe the effect of class variable imbalance due to device movement under the edge network, in the first 50 rounds of training, the number of classes of the training data is 10, as well as the class distribution is balanced. Starting from the 50th round, a new class is added or reduced in the 50th, 100th, 150th, 200th, and 250th rounds, respectively, as well as the class distribution of their data also changes randomly. In the class increase imbalance, the number of categories changes from 10 to 11, 12, 13, 14, and 15 at rounds 50, 100, 150, 200, and 250; in the class decrease imbalance, the number of categories changes from 10 to 9, 8, 7, 6, and 5 at rounds 50, 100, 150, 200, and 250.

Due to the complexity of image data in the ImageNet dataset, we choose the AlexNet image classification model as the backbone model. AlexNet is the model that first introduced convolutional neural network into the field of computer vision and achieved a breakthrough, winning the ILSVRC 2012 championship. AlexNet is a convolutional network consisting of five convolutional layers and three fully connected layers, with a total of eight layers of convolutional neural network. We tuned the parameters of AlexNet according to the hardware configuration of the physical machine and the characteristics of the dataset.

### 4.1.2. Edge Network Setup

To evaluate the scheme proposed in this paper and compare it with other schemes in the literature, we built a real distributed training edge network and deployed it on a physical device, as shown in Figure 1. We use a Raspberry Pi 4B (Raspberry Pi 4B) as the edge device; a workstation with a GPU as the edge server, which is closer to the edge device; and a server superior to the workstation GPU as the aggregation server, which is further away from the edge device than the edge server. The specific hardware configuration is shown in Table 1, which is common in actual production environments and has some generality.

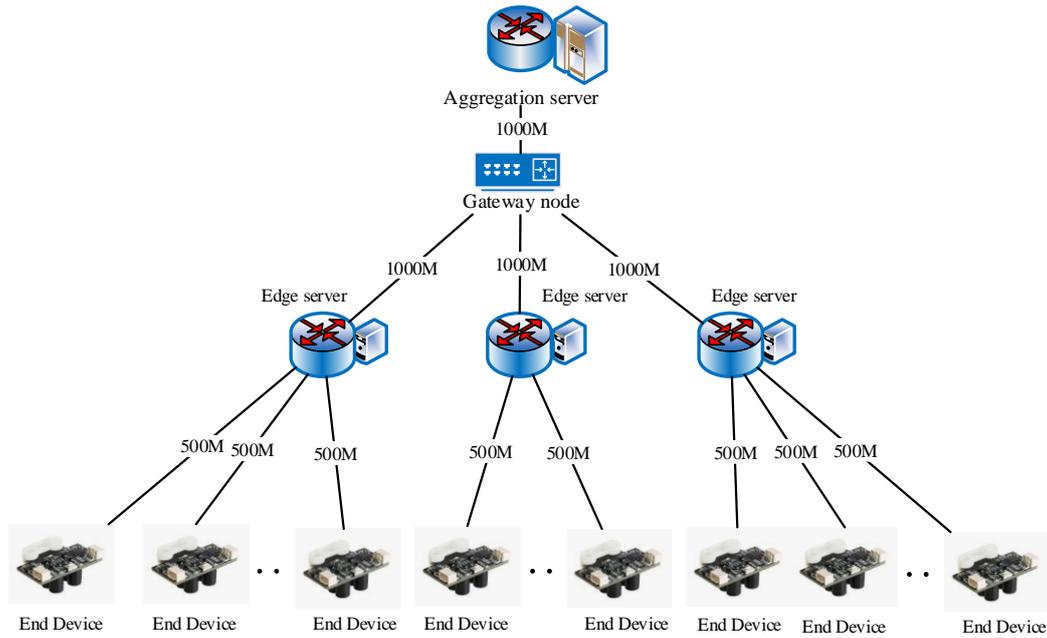

Figure 1 Distributed training edge network topology diagram

Table 1 Hardware configuration information of each edge device

| Node Type | End Device | | Edge Server | | Aggregation Server | |
|---|---|---|---|---|---|---|
| OS | Raspberry Pi OS | | Ubuntu 20.04 LTS | | Ubuntu 16.04 LTS | |
| CPU | ARM CortexA-72 @ 1.5GHz | | Intel(R) Core(TM) i5-10500 @3.10GHz | | Intel(R) Xeon(R) Gold 6130 CPU @2.10Ghz | |
| Memory | 2GB DDR4 | | 16GB DDR4 | | 512GB DDR4 | |
| GPU | 500 MHZ VideoCore VI | | NVIDIA GeFroce RTX 2060 SUPER | | NVIDIA Tesla P100 *4 | |
| Wireless Network | 802.11ac(2.4/5GHz) Bluetooth 5.0 | | 802.11ac(2.4/5GHz) Bluetooth 5.0 | | 802.11ac(2.4/5GHz) Bluetooth 5.0 | |
| Wired Network | Gigabit Ethernet | | Gigabit Ethernet | | Gigabit Ethernet | |
| Power supply | Idle power | 10W | Idle power | 110W | Idle power | 500W |
| | CPU full load power | 15W | CPU full load power | 65W | CPU full load power | 125W |
| | GPU Full Load Power | 12W | GPU Full Load Power | 180W | GPU Full Load Power | 2400W |

### 4.1.3. Comparison Scheme

In order to compare the effectiveness of FCVI, we implemented the scenarios proposed in this paper and the most representative comparison scenarios. All these scenarios are implemented on our testbed using the same dataset and backbone model, and the details are described as follows.

(1) FCVI: FCVI's are set according to the algorithm above.
(2) FedRL[21]: FedRL can infer the composition of the training data for each round of FL through the aggregation server and mitigate the effect of category imbalance by a new loss function, Ratio Loss.

(3) K-SMOTE[20]: K-SMOTE updates its local model by directly interacting with the connected neighboring edge servers and generates synthetic data for a few classes based on linear interpolation to rebalance the local token set on the edge servers. In K-SMOTE, the aggregation server is not involved in the training process except for the initial model deployment.

(4) FedAvg [24]: In FedAvg, all edge servers use asynchronous stochastic gradient descent (SGD) in parallel to compute and update their weights, and then the server collects updates from clients and aggregates them using the FedAvg algorithm.

### 4.1.4. Evaluation Metrics

*Accuracy* is the ratio of the number of all correct classifications of the classifier to the total number of.

$$Accuracy = \frac{TP + TN}{TP + FP + TN + FN}$$

Where, $TP$、 $FP$、 $FN$ and $TN$ are abbreviated forms of true positive, false positive, false negative and true negative, respectively.

Precision is the ratio of all "correctly classified numbers ($TP$)" to all "actually classified numbers ($TP + FP$)".

$$Precision = \frac{TP}{TP + FP}$$

Recall is the ratio of all "number of correct classifications ($TP$)" to all "actual classifications as correct ($TP + FN$)".

$$Recall = \frac{TP}{TP + FN}$$

F1 is the summed average of the precision rate and recall rate.

$$F1 = \frac{2Precision * Recall}{Precision + Recall} = \frac{2TP}{2TP + FP + FN}$$

### 4.2. Analysis of Experimental Results

In this section, the effectiveness of this paper's scheme and the model performance is experimentally evaluated. Firstly, it focuses on comparing the trends of accuracy rates of various methods to verify the ability of this paper's scheme to monitor class variable imbalance. Secondly, the accuracy, recall, precision and F1 values of the compared solutions are compared under different rounds of class variable imbalance to verify that this solution can effectively mitigate the degradation of model performance due to class variable imbalance. The experimental results in this section are all the expected values of 10 independent experiments.

### 4.2.1. Monitoring Method Validity

In this section, FCVI is compared with FedRL, K-SMOTE, and FedAvg to assess the ability of each method to monitor class variable imbalance. Specifically, class variable imbalance can be divided into two cases, class increasing imbalance and class decreasing imbalance, and this section compares the accuracy variation curves of these strategies under these two classes of variable imbalance, respectively.

Figure 2 shows the comparison of the accuracy change curves under class-increasing imbalance. After starting from the 50th round, the number of classes of the classification model will increase by one class every 50 rounds. When the number of classes increases, the accuracy of K-SMOTE and FedAvg decreases, and the accuracy of both FCVI and FedRL increases. In contrast, the accuracy curve of FCVI has a significant increase and is significantly higher than the other three methods. Figure 3 shows the comparison of the accuracy change curves under the class reduction imbalance. After starting from the 50th round, the number of classes of the classification model will decrease by one class every 50 rounds. When the number of classes is reduced, the accuracy of FedRL, K-SMOTE and FedAvg obviously decreases, while the accuracy of FCVI only slightly decreases and remains basically stable overall. This is because K-SMOTE and FedAvg are unable to process the increased and decreased classes, which affects the overall accuracy change. FedRL is able to process the increased and decreased classes, but does not monitor the change in the number of classes in a timely manner, resulting in the collation accuracy being affected. On the contrary, FECI can accurately monitor the changes in the category data by passing the training gradient parameters and timely expand them using unlabeled data, which ultimately has little impact on the collation accuracy and ensures the accuracy of the model.

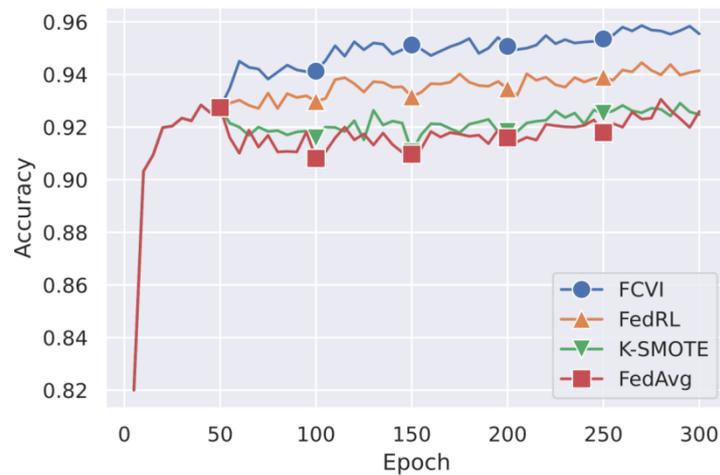

Figure 2 Comparison of accuracy change curves under class increasing imbalance.

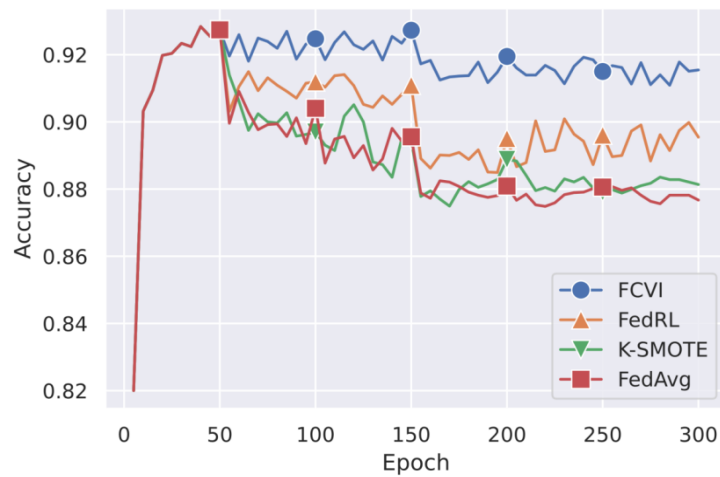

Figure 3 Comparison of accuracy change curves under class decreasing imbalance.

**4.2.2. Model Performance Comparison**

To evaluate the execution efficiency of FCVI, this section compares FCVI with FedRL, K-SMOTE, and FedAvg for different model training rounds under class-variable imbalance. In the first 50 rounds of training, the number of classes and class distribution of the training data does not change, and from the 50th round onward, the number of new or reduced classes and other classes are also randomly changed in the 50th, 100th, 150th, 200th, and 250th rounds, respectively. In this paper, we will compare the number of classes and class distribution of the training data after the occurrence of changes in the 50th round, the 100th, 150th, 200th, 250th, and 300th rounds, respectively. 200, 250, and 300 rounds of model performance comparison.

Table 2 Comparison of accuracy, recall, precision and F1 values for differente poch under class increasing imbalance

| Epoch | | 100 | 150 | 200 | 250 | 300 | Average | △ |
|---|---|---|---|---|---|---|---|---|
| Accuracy | FCVI | **0.941** | **0.951** | **0.951** | **0.954** | **0.955** | **0.950** | — |
| | FedRL | 0.930 | 0.932 | 0.935 | 0.939 | 0.941 | 0.935 | ⇑0.015 |
| | K-SMOTE | 0.916 | 0.911 | 0.918 | 0.925 | 0.925 | 0.919 | ⇑0.031 |
| | FedAvg | 0.908 | 0.910 | 0.916 | 0.918 | 0.926 | 0.916 | ⇑0.035 |
| Recall | FCVI | **0.625** | **0.658** | **0.618** | **0.592** | **0.609** | **0.620** | — |
| | FedRL | 0.618 | 0.606 | 0.583 | 0.562 | 0.548 | 0.583 | ⇑0.037 |
| | K-SMOTE | 0.544 | 0.510 | 0.467 | 0.468 | 0.442 | 0.486 | ⇑0.134 |
| | FedAvg | 0.519 | 0.479 | 0.451 | 0.448 | 0.419 | 0.463 | ⇑0.157 |
| Precision | FCVI | **0.640** | **0.666** | **0.653** | **0.629** | **0.629** | **0.643** | — |
| | FedRL | 0.620 | 0.615 | 0.592 | 0.574 | 0.550 | 0.590 | ⇑0.053 |
| | K-SMOTE | 0.567 | 0.543 | 0.492 | 0.483 | 0.474 | 0.512 | ⇑0.132 |
| | FedAvg | 0.702 | 0.512 | 0.473 | 0.459 | 0.448 | 0.519 | ⇑0.125 |
| F1 | FCVI | **0.624** | **0.657** | **0.622** | **0.596** | **0.610** | **0.622** | — |
| | FedRL | 0.611 | 0.602 | 0.579 | 0.560 | 0.531 | 0.577 | ⇑0.045 |
| | K-SMOTE | 0.525 | 0.492 | 0.437 | 0.442 | 0.416 | 0.462 | ⇑0.159 |
| | FedAvg | 0.499 | 0.449 | 0.413 | 0.416 | 0.385 | 0.432 | ⇑0.190 |

Tables 2 and 3 show the four evaluation results of the classification model under different rounds of the method under class increasing imbalance and class decreasing imbalance, respectively, showing the model performance after 50 rounds of occurrence of changes in the number of classes, where '△' indicates the improvement of the method FCVI in this paper compared with other compared methods. It is observed that in class increase imbalance, the method in this paper outperforms other comparative methods in different training rounds, as well as in average accuracy, average recall, average precision, and F1 values of 0.029-0.0157, 0.037-0.125, and 0.030-0.190. Similarly, in class reduction imbalance, the method in this paper outperforms other comparison methods in different training rounds as well as has, 0.013-0.0076, 0.013-0.058, and 0.016-0.100 advantages in average accuracy, average recall, average precision, and F1 values.

Table 3 Comparison of accuracy, recall, precision and F1 values for different epoch underclass decreasing imbalance

| Epoch | | 100 | 150 | 200 | 250 | 300 | Average | Δ |
|-------|------|------|------|------|------|------|---------|---|
| Accuracy | FCVI | **0.925** | **0.923** | **0.926** | **0.920** | **0.915** | **0.922** | — |
| | FedRL | 0.912 | 0.911 | 0.895 | 0.896 | 0.895 | 0.902 | ⇑0.020 |
| | K-SMOTE | 0.897 | 0.895 | 0.889 | 0.879 | 0.881 | 0.888 | ⇑0.034 |
| | FedAvg | 0.904 | 0.896 | 0.881 | 0.881 | 0.877 | 0.888 | ⇑0.034 |
| Recall | FCVI | **0.629** | **0.662** | **0.646** | **0.709** | **0.718** | **0.673** | — |
| | FedRL | 0.600 | 0.635 | 0.622 | 0.692 | 0.730 | 0.656 | ⇑0.017 |
| | K-SMOTE | 0.540 | 0.580 | 0.604 | 0.627 | 0.693 | 0.609 | ⇑0.064 |
| | FedAvg | 0.533 | 0.561 | 0.574 | 0.634 | 0.684 | 0.597 | ⇑0.076 |
| Precision | FCVI | **0.631** | **0.673** | **0.647** | **0.710** | **0.722** | **0.676** | — |
| | FedRL | 0.608 | 0.644 | 0.620 | 0.695 | 0.740 | 0.661 | ⇑0.015 |
| | K-SMOTE | 0.567 | 0.617 | 0.611 | 0.645 | 0.699 | 0.628 | ⇑0.049 |
| | FedAvg | 0.561 | 0.595 | 0.597 | 0.652 | 0.686 | 0.618 | ⇑0.058 |
| F1 | FCVI | **0.630** | **0.662** | **0.643** | **0.707** | **0.718** | **0.672** | — |
| | FedRL | 0.596 | 0.628 | 0.614 | 0.689 | 0.728 | 0.651 | ⇑0.021 |
| | K-SMOTE | 0.516 | 0.561 | 0.584 | 0.607 | 0.678 | 0.589 | ⇑0.083 |
| | FedAvg | 0.500 | 0.536 | 0.547 | 0.612 | 0.662 | 0.572 | ⇑0.100 |

Based on the analysis of the above results, this verifies the effectiveness of this paper's method in alleviating the class-variable imbalance problem. It can be concluded that the method in this paper can effectively improve the performance of the model by aggregating servers and edge servers under FL gradient detection method and class variable learning algorithm so that the model can learn new class data and prevent forgetting old class data during the training process.

## 5. CONCLUSIONS

We proposed the Federated Semi-supervised Learning for Class Variable Imbalance (FCVI) method for solving the class variable imbalance problem in federation learning. Compared to other methods, FCVI does not require the collection of additional model parameters while mitigating the negative impact of class variable imbalance on model training. This work aims to mitigate the data imbalance problem caused by the changing number of classes through the federal gradient monitoring method and the class-variable learning algorithm. First the federal gradient monitoring method is used to monitor and locate the proportion of category changes, and then the data imbalance due to the number of category changes is mitigated by monitoring the resulting proportion of category changes and the class-variable learning algorithm. Experiments show that FCVI can guarantee the performance and execution efficiency of distributed training of deep learning models while adapting to the increase and decrease of the number of categories.

## AUTHORS

**Zehui Dong** received North University of China, and he did Engineering Degree in Software Engineering. Currently, he is pursuing his Master's degree in Software Engineering from Inner Mongolia University of Technology. His research interests include edge computing, federated learning.

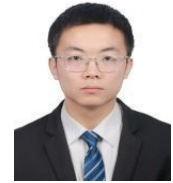

**Wenjing Liu** received the B.S. degree from Inner Mongolia University of Technology, Hohhot, China, in 2011, the M.S. degree from Beijing University of Posts and Telecommunications, Beijing, China, in 2018. She is a college lecturer in Inner Mongolia University of Technology. Her research interests include algorithm design and assessment and artificial intelligence.

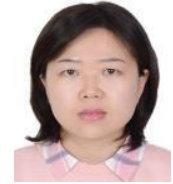

**Siyuan Liu** received the B.S. degree in 2021 from Tianjin Gongye University, Tianjin, China. She is currently pursuing the M.S. degree in Inner Mongolia University, Hohhot, China. Her research interests include edge computing and big data technology.

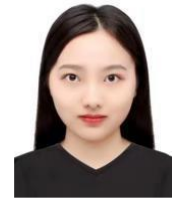

**Xingzhi Chen** graduated from Chongqing Institute of Technology with a degree in Software Engineering. He is currently studying for a master's degree in software engineering at Inner Mongolia University of Technology. His research interest covers pattern recognition and single image rain removal.

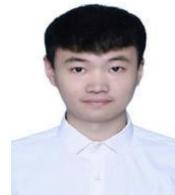